\pdfoutput=1

\documentclass[11pt]{article}

\usepackage[]{acl}

\usepackage{times}
\usepackage{latexsym}
\usepackage[T1]{fontenc}
\usepackage[utf8]{inputenc}
\usepackage{microtype}
\usepackage{inconsolata}
\usepackage{graphicx}
\usepackage{url}
\usepackage{times}
\usepackage{latexsym}
\usepackage{longtable}
\usepackage{booktabs}
\usepackage{enumitem}
\usepackage{multirow}
\usepackage{tikz}
\usepackage{amsmath}
\usepackage[ruled, noend]{algorithm2e} 
\usepackage{tabularray}
\usepackage{centernot}
\usepackage{placeins}
\usepackage{amsfonts}

\usepackage{booktabs}
\usepackage{tcolorbox}
\usepackage{listings}
\definecolor{dkgreen}{rgb}{0,0.6,0}
\definecolor{gray}{rgb}{0.5,0.5,0.5}
\definecolor{mauve}{rgb}{0.58,0,0.82}
\newcommand{\sref}[1]{\S\ref{#1}}

\title{Adversarial Math Word Problem Generation}

\author{\\ \bf
    Roy Xie \hspace{4mm}
    Chengxuan Huang \hspace{4mm}
    Junlin Wang \hspace{4mm}
    Bhuwan Dhingra \\ [1ex]
    Duke University \\ [1ex]
    \texttt{\{ruoyu.xie, jonathan.huang, junlin.wang2\}@duke.edu}\\
    \texttt{\{bdhingra\}@cs.duke.edu} 
    \\
    }

\begin{document}
\maketitle

\begin{abstract}
Large language models (LLMs) have significantly transformed the educational landscape. As current plagiarism detection tools struggle to keep pace with LLMs' rapid advancements, the educational community faces the challenge of assessing students' true problem-solving abilities in the presence of LLMs.
In this work, we explore a new paradigm for ensuring fair evaluation---generating
adversarial examples which preserve the 
structure and difficulty of the original questions aimed
for assessment, but are unsolvable by LLMs. Focusing on the domain of math word problems, we leverage abstract syntax trees to structurally generate adversarial examples that cause LLMs to produce incorrect answers by simply editing the numeric values in the problems. We conduct experiments on various open- and closed-source LLMs, quantitatively and qualitatively demonstrating that our method significantly degrades their math problem-solving ability. We identify shared vulnerabilities among LLMs and propose a cost-effective approach to attack high-cost models. Additionally, we conduct automatic analysis to investigate the cause of failure, providing further insights into the limitations of LLMs.
\footnote{Data and code are available: \url{https://github.com/ruoyuxie/adversarial_mwps_generation}} 
\end{abstract}
\section{Introduction}
Recent advances in large language models (LLMs) have revolutionized the world of education, primarily due to the great improvements in their natural language generation and problem-solving capabilities. This has transformed how students access information and complete assignments, raising significant concerns among educators in accurately evaluating students' true problem-solving abilities with the presence of such powerful tools \citep{OpenAI2023GPT4TR, kung2023performance, callanan2023can}. While efforts like plagiarism detection exist \citep{kirchenbauer2023watermark, mitchell2023detectgpt}, their effectiveness in identifying LLM-generated content is limited \citep{liang2023gpt, chaka2023detecting}, underscoring the need for more advanced anti-plagiarism methods to match LLM advancements. 

At the same time, adversarial attacks on LLMs have gained more attention due to increased awareness of the potential risks associated with LLMs. Most work on adversarial attacks focuses on developing prompts to elicit specific outputs from LLMs \citep{zhang2020adversarial, zou2023universal,carlini2023aligned}
Recent work suggests that even the most powerful aligned LLMs are vulnerable to such attacks \citep{zou2023universal, carlini2023aligned}. 
Hence, we might expect that as LLMs become stronger,
detecting their outputs will become more difficult,
but adversarial examples may still persist \citep{ilyas2019adversarial,wei2023jailbroken}.

To this end, we introduce a new paradigm for generating homework assignments that LLMs cannot solve by utilizing adversarial attacks. We focus on math word problems (MWPs), which present a unique and challenging intersection of language and mathematics. In this work, we ask the question: ``\textit{Can LLMs still solve an MWP after changing its numeric values?}''
We aim to create MWPs that LLMs are unable to solve while \textit{maintaining the original difficulty and coherence of the problems}. Our aim is not just to challenge LLMs but to do so in a way that reflects real-world educational standards to ensure the MWPs remain educationally valuable and relevant. 

\begin{figure*}[htpb!]
    \centering
    \includegraphics[width=\textwidth]{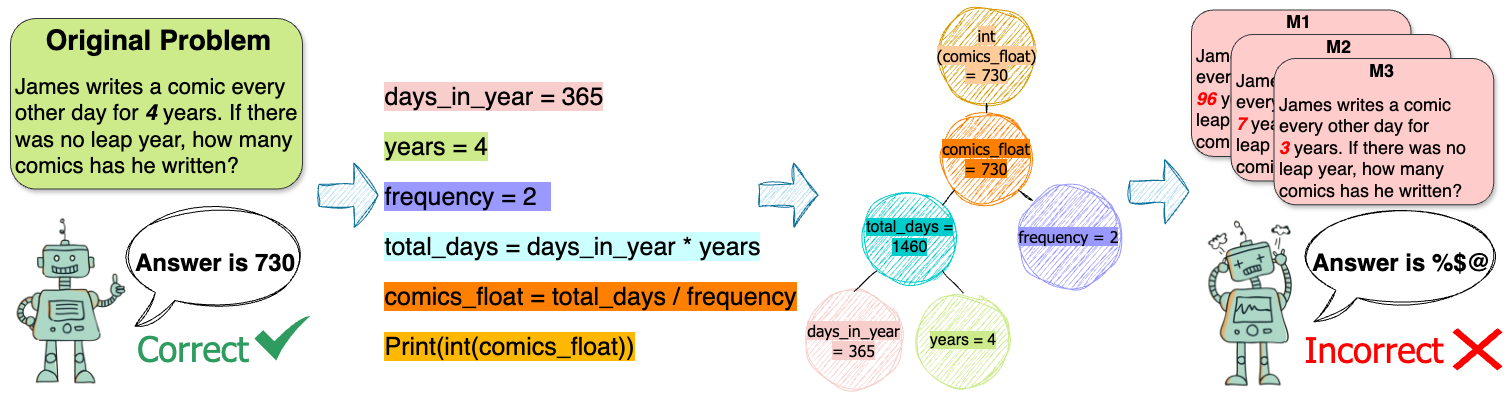}
    \caption{\textbf{Method Overview:} Given a MWP that an LLM can correctly solve, our method first transforms it into Python code. The Python code then is converted into an AST representation, which is used to generate adversarial problems by modifying the numeric values in a controllable manner.
    We place constraints on the nodes of the AST to ensure that the modified problem maintains the same difficulty level as the original problem.
    Despite this, we find that the resulting adversarial examples cause LLMs to predict incorrect answers.
    }
    \label{fig:method_overview}
    \vspace{-1em}
\end{figure*}

While conceptually simple, doing this automatically is challenging as directly modifying numbers in the problem could lead to nonsensical problems. For example, consider the problem ``A class has 6 male students and half as many female students. How many students in total?'' Directly change the number of male students from 6 to 5 (or any odd numbers) without checking its intermediate computational steps might result in ``2.5 female students,'' which is not only illogical but also introducing fraction to the problem, which might change the indented difficulty. Similarly, changing the number from 6 to 624 might make the problem unrealistic and much harder for students. To generate meaningful and coherent problem variations, it is essential to consider the logical implications of number modifications and ensure that the resulting problem remains plausible and solvable. We further discuss the importance of educational constraints in \sref{educational_context}.

To effectively generate modifications in large scale to assess the robustness of LLMs also requires: (i) The answer to the altered problem changes must be recomputed automatically; and (ii) Preserving the difficulty and validity of the modified problem requires us to ensure that all the intermediate calculations in the new problem are consistent with the original problem. To tackle these challenges, we first convert MWPs to a code representation and leverage abstract syntax trees (ASTs) to map each calculation step into a node. We then define educational constraints for each node to ensure all the desired properties are preserved for the generated new problem. 

In this work, we evaluate several LLMs and demonstrate the effectiveness of our method by achieving a significant attack success rate (ASR). Our approach outperforms the previous rephrasing attack by an average of 62\% ASR. We investigate universal attacks and attack transferability, proposing a cost-effective approach to attack high-cost models (e.g., GPT-4) by reducing API request calls by 90\% while achieving high performance. We conduct human evaluations to verify that our generated problems indeed preserve the original coherence and difficulty. Furthermore, we perform a regression analysis and find that our adversarial examples exploit different weaknesses of each model, offering valuable insights into LLM's limitation.

\section{Background and Related Work}

\paragraph{Fair Evaluation for Educational Purpose}
As LLMs become more adept at generating human-like text, it becomes increasingly difficult to distinguish between student-and machine-generated content \citep{chaka2023detecting,liang2023gpt}, which poses significant challenges for educational institutions in ensuring fair evaluation of student work \citep{yan2024practical}. This issue also extends beyond traditional written assignments, as LLMs can now provide detailed solutions to complex problems across various disciplines \citep{abedi2023beyond}. Consequently, educators must develop new strategies to assess student understanding and maintain the integrity of the evaluation process \citep{liu2023future}.

\paragraph{LLMs Math-solving Ability}
Our work also closely relates to LLM's math reasoning ability \citep{yu2023metamath, xu2023wizardlm}. Studies found that LLMs can significantly improve their math-solving ability through prompt engineering \citep{yu2023metamath,DBLP:conf/acl/ImaniD023}, such as chain-of-thought (CoT) \citep{wei2022chain}. Converting the MWPs' solving steps into symbolic representations can also improves LLM's performance \citep{li2023chain,he2023solving,gao2023pal}.\footnote{While different prompting strategies exist, our focus is to generate adversarial examples that LLMs are unable to solve. Therefore, we try to minimize prompt engineering by using a unified prompt with zero-shot CoT for all evaluations. The prompt can be found in the appendix \ref{appendix:main_prompt}.}

\paragraph{Adversarial Attacks on MWPs} 
Adversarial attacks on LLMs involve modifying prompts to alter their behavior \citep{zhang2020adversarial, zou2023universal, carlini2023aligned}. In this work, we modify MWPs to cause LLMs to output incorrect answers. \citet{bubeck2023sparks} conducted limited memorization tests on MWPs by randomly changing numeric values, suggesting that state-of-the-art LLMs do not solely rely on memorization but apply general solution methods. However, their study's sample size was relatively small, and we demonstrate that modifying numbers in MWPs causes LLMs to fail on a larger scale. On the other hand, \citet{zhou2023mathattack} rephrased MWPs by changing words while preserving numeric values. This approach risks altering the original context and introducing inconsistencies and problematic content (see Appendix \ref{appendix:rephrasing_example}), requiring extensive human validation. Therefore, we focus on altering the numerical values in this work and preserve the underlying logic. 
\section{Methodology}\label{sec:generation_method}
MWPs are presented in natural language, which creates challenges for systematic structural and syntactic modifications. In this section, we describe our approach to generating adversarial MWPs. An overview of our method can be found in Figure \ref{fig:method_overview}. We denote an original problem-answer pair as $p = (x, y)$, where $x$ is a sequence of tokens $x_{1}, \dots, x_{n}$ in the problem and $y$ is its ground-truth answer. We define $G$ as a ground-truth function that computes the correct answer for a math problem 
and $F$ as a function which maps the elements of a sequence as:
$F(\tilde{x}_i) \sim \mathbb{R}$ (i.e., a random real number) if 
$x_i$ is numeric and $F(\tilde{x_i}) = x_i$ otherwise.\footnote{In our case $G$ is generated Python code.} We denote the set of all possible adversarial modifications to it as:
\begin{equation}
    A(x, y) = \{(\tilde{x}, \tilde{y}): \tilde{x}_i \in F(x), \tilde{y} = G(\tilde{x})\}.
\end{equation}

$A(x, y)$ will also consist of many
unnatural and difficult problems,
hence later in \sref{sec:node_constraints} we will introduce \textit{filtering constraints}
for selecting the adversarial examples that preserve difficulty.

\subsection{Mapping From MWPs to Code to Tree} 
To structurally modify MWPs, we first use GPT-4 to generate the Python code that reflects the solution steps given the problem and its final answer.\footnote{This is an one-time operation. We only require the problem and its final answer, which is almost always given in the context of math problems. The prompt is in the Appendix \ref{appendix:code_gen_prompt}.} 
Next, we utilize the AST, a tree structure for programming language code, to convert the generated Python code into a tree representation for controllable new problem generation. We build the ASTs by traversing through the Python code and constructing nodes corresponding to each statement. The final print statement that outputs the answer is the root of the tree. 
Each AST has mainly two types of nodes: \textit{operation nodes}, which carry out the operations in the tree and are the non-leaf nodes, and \textit{variable nodes}, which correspond to the numeric values from the problem and are the leaf nodes. We discuss the nodes in ASTs in detail in Appendix \ref{appendix:node_detail} and conduct analysis on the quality of generated Python code and ASTs in \sref{sec:python_code_analysis}.

\subsection{Adversarial Example Generation}
In an adversarial setting, attackers can perform multiple rounds of attack on the system, and any successful attack signifies the possible flaws of the system \citep{carlini2017towards}. Similarly, we generate adversarial examples which are different versions of the original MWPs to attack LLMs.

\paragraph{Educational Context}\label{educational_context}
It's essential to maintain the original \textit{difficulty} and \textit{coherence} of the MWPs despite changes in their numeric values. For instance, changing one-digit multiplication to four-digit multiplication significantly alters the problem's complexity. To maintain mathematical logic intact, the order of magnitude of numbers in the problem should also remain unchanged. For example, changing a problem from ``Jack ate 2 out of a total of 5 apples'' to ``Jack ate 5 out of a total of 2 apples'' not only introduces the concept of negative numbers but also alters the problem's logical structure and coherency. Such modifications can create unrealistic scenarios, potentially confusing students and leading to ineffective learning experiences \cite{vilenius2008association}.

\paragraph{Filtering Constraints}\label{sec:node_constraints} 
To minimize the difference between the original MWP and its adversarial counterpart while adhering to the educational context, we define a set of Boolean constraints for each node in the AST to control the generation quality. Given the original node value $h$ and its new value $h'$, a newly generated problem is valid if and only if $h'$ satisfies the same set of constraints that $h$ has, for all nodes. A list of node constraints is: 
\begin{itemize} [leftmargin=*,noitemsep,nolistsep]
\setlength{\itemsep}{1pt}
\setlength{\parskip}{1pt}
\setlength{\parsep}{1pt}
\item \textbf{Positivity}: if $h$ is positive, then $h'$ should be positive. This constraint avoids generating phrases like ``get 5 apples from 2'' since this would produce a negative node.
\item \textbf{Integer}: if $h$ is an integer, then $h'$ should remain an integer. This ensures that phrases like ``the first half of 3 people'' wouldn't appear, since the original problem likely has an integer value in the intermediate node. 
\item \textbf{Proper Fraction}: if $h$ is between 0 and 1, then $h'$ should be between 0 and 1. This prevents phrases such as ``John eats 4/3 of his chimichanga'' and ``the bucket is filled till 150\% full'' from being produced since the generated problems wouldn't make logical sense for many problems when a number is no longer a proper fraction. 
\end{itemize}

\paragraph{Constrictive Generation Methods}
The node constraints ensure the adversarial examples are valid in the numerical sense; instances that don't make logical sense like ``$28$-hour work day'', which is not common, can still be generated. Different values for the variable nodes can also lead to vastly different difficulty levels, as $h'$ can be significantly larger than $h$. To mitigate this, we propose three generation methods to distinguish generated adversarial examples with different levels of difficulty. We decompose the values $h\approx a\times 10^b$, where $a$ is an integer that is not divisible by 10 and with at most $c$ digits. Depending on the numbers $h$, $a$, and $b$, the generation methods are defined as follows: 

\begin{itemize} [leftmargin=*,noitemsep,nolistsep]
\setlength{\itemsep}{1pt}
\setlength{\parskip}{1pt}
\setlength{\parsep}{1pt}
    \item \textbf{$M1$ Free Generation}: Allowing the generated problem to have a wide range of numbers with minimal constraints, with each $h'$ to be $a'\times 10^b$. Regardless of the value of $a$, $a'$ is drawn from a uniform distribution between $1$ and $10^c$. If the variable node $v$ is a divisor of a division node, then $a'$ is drawn from a uniform distribution between $1$ and $\left\lfloor10^{c/2}\right\rfloor$. 
    \item \textbf{$M2$ Count of Digits}: Constraining the generated value $h'$ has the same number of digits as $h$. For example, for $h=100$, some possible $h'$ are 942, 589, or 264. This ensures $h'$ is always within a reasonable range relative to $h$ and maintains similar problem difficulty levels. For more variability, $h'$ can range from 1 to 99 if $h$ has only one digit. For decimal variable values like $h=1.25$, we use $a=125$ to generate numbers like $a'=473$, then convert them back to $h'=4.73$.
    \item \textbf{$M3$ Count of Scientific Numbers}: Constraining the new value shares a similar scientific digit count and is within a similar range to the original value. For example, for $h=1500$, $h'\sim \text{Pois}(h)$ can be 1700, 800, 1200, where \text{Pois} is the Poisson distribution. The rationale is that \textit{larger numbers do not result in more difficult problems}, but more scientific numbers do. An example would be comparing the numbers 150,000 and 172,568.
\end{itemize}
$M3$ is the most restrictive generation, followed by $M2$, and $M1$ is the least restrictive:
\begin{equation}
M1(A(x, y)) \supseteq M2(A(x, y)) \supseteq M3(A(x, y)).
\end{equation}
The more restrictive a method is, the closer the difficulty levels and coherence remain between an adversarial example and its original version. Thus, we use $M3$ as our main generation method since it ensures that adversarial examples adhere to all constraints, maintaining original difficulty and coherence. While $M1$ and $M2$ generations may not fit the educational context, we aim to study LLMs' math-solving abilities by simply altering numeric values. We discuss how different methods impact model performance in \sref{sec:model_performance}. We present detailed descriptions of each generation method in Appendix \ref{appendix:gen_detail} and their generated examples in Table \ref{appendix:problem_sample}.
 
\section{Experiments and Results}
In this section, we present multiple experiments
and demonstrate the effectiveness of our method on attacking various LLMs.

\subsection{Experimental Setup}
\paragraph{Datasets}\label{sec:dataset}
We generate problem variants from GSM8K \citep{cobbe2021training} and MultiArith \citep{roy-roth-2015-solving}. Both datasets are commonly used for evaluating LLMs' mathematical capabilities, and MultiArith is sourced directly from math worksheets used by elementary school students to practice math problems \citep{roy2015reasoning}.\footnote{While MWPs exhibit different levels of difficulty, we focus on MWPs at this particular difficulty level as a proof of concept in this work.}

\begin{table*}[ht!]
\centering
\scriptsize
\begin{tabular}{@{}lr|rr|rr|rr|r|rr|rr|rr@{}}
\toprule
& \multicolumn{1}{c}{\textbf{}} & \multicolumn{6}{c}{\textbf{MultiArith}} & \multicolumn{7}{c}{\textbf{GSM8K}} \\
& \multicolumn{1}{c}{} & \multicolumn{2}{c}{\textbf{$M3$}} & \multicolumn{2}{c}{\textbf{$M2$}} & \multicolumn{2}{c|}{\textbf{$M1$}} & \multicolumn{1}{c}{} & \multicolumn{2}{c}{\textbf{$M3$}} & \multicolumn{2}{c}{\textbf{$M2$}} & \multicolumn{2}{c}{\textbf{$M1$}} \\
\textbf{Model} & OA & AA & ASR & AA & ASR & AA & ASR & OA & AA & ASR & AA & ASR & AA & ASR \\
\midrule
Mistral 7B& 37.0 & 0.0 & 100.0 & 0.0 & 100.0 & 0.0 & 100.0 & 29.0 & 0.0 & 100.0 & 0.0 & 100.0 & 0.0 & 100.0 \\
MetaMath 7B & 100.0 & 74.0 & 26.0 & 10.0 & 90.0 & 0.0 & 100.0 & 95.0 & 28.0 & 71.0 & 9.0 & 91.0 & 0.0 & 100.0 \\
Llama 3 8B& 17.0 & 0.0 & 100.0 & 0.0 & 100.0 & 0.0 & 100.0 & 21.0 & 0.0 & 100.0 & 0.0 & 100.0 & 0.0 & 100.0 \\
Llama 2 13B& 12.0 & 0.0 & 100.0 & 0.0 & 100.0 & 0.0 & 100.0 & 10.0 & 0.0 & 100.0 & 0.0 & 100.0 & 0.0 & 100.0 \\
WizardMath 13B& 89.0 & 20.0 & 78.0 & 5.0 & 94.0 & 0.0 & 100.0 & 89.0 & 11.0 & 88.0 & 2.0 & 98.0 & 0.0 & 100.0 \\
Vicuna 13B& 76.0 & 4.0 & 95.0 & 1.0 & 99.0 & 0.0 & 100.0 & 60.0 & 0.0 & 100.0 & 0.0 & 100.0 & 0.0 & 100.0 \\
CodeLlama 34B& 11.0 & 0.0 & 100.0 & 0.0 & 100.0 & 0.0 & 100.0 & 6.0 & 0.0 & 100.0 & 0.0 & 100.0 & 0.0 & 100.0 \\
MetaMath 70B& 99.0 & 86.0 & 13.0 & 30.0 & 70.0 & 0.0 & 100.0 & 98.0 & 50.0 & 49.0 & 17.0 & 83.0 & 0.0 & 100.0 \\
GPT-3.5 & 97.0 & 74.0 & 24.0 & 47.0 & 52.0 & 0.0 & 100.0 & 91.0 & 52.0 & 43.0 & 31.0 & 66.0 & 0.0 & 100.0 \\
\midrule
Average & 60.0 & 28.7 & 70.7 & 10.3 & 78.3 & 0.0 & 100.0 & 55.4 & 15.9 & 83.4 & 6.6 & 93.1 & 0.0 & 100.0 \\
\bottomrule
\end{tabular}
\caption{\textbf{Main Attacks:} Performance of three different generation methods. Simply changing numeric values consistently cause performance drop across all LLMs, even with the most restrictive generation method.}
\label{table:main_results}
\vspace{-2em}
\end{table*}

\paragraph{Models}
We conduct experiments on the following open-source models: MetaMath 7B, 70B \citep{yu2023metamath}, Mistral 7B \citep{jiang2023mistral}, 
Llama 3 8B \citep{llama3modelcard},
Llama 2 13B \citep{touvron2023llama}, WizardMath 13B \citep{xu2023wizardlm}, Vicuna 13B \citep{chiang2023vicuna}, and CodeLlama \citep{roziere2023code}. Additional, we evaluate two closed-source models: GPT-4 \citep{OpenAI2023GPT4TR} and GPT-3.5 \citep{chatgpt}.\footnote{GPT-4-0125-preview and GPT-3.5-turbo-0125.} The model selection was largely based on the LLMs' math-solving ability. Our intention is to evaluate a wide range of LLM performances, including the math-tuned LLMs such as MetaMath and WizardMath, popular open-source LLMs such as Llama 3 and Mistral, and the API-based GPT models.

\paragraph{Metrics}
We follow the previous adversarial attack literature to measure the attacks, which looks for at least one perturbation that fools the model \citep{croce2020robustbench,pruthi2019combating,jia2017adversarial}. A problem is considered incorrect if it has at least one incorrect variation, also known as an \textit{adversarial example}.  Given a set of original problem-answer pairs $P$ and a LLM $L$, we define:\footnote{OA, AA, ASR are reported in \% in this work.}

\begin{itemize}[leftmargin=*,noitemsep,nolistsep]
  \setlength{\itemsep}{0pt}
  \setlength{\parskip}{0pt}
  \setlength{\parsep}{0pt}
\item \textbf{Original Accuracy (OA)}: the accuracy of \(L\) on the original problems,
\begin{equation}
\text{OA}(L) = \frac{\sum_{(x, y) \in P} \mathrm{1}\{L(x) = y\}}{|P|}.
\end{equation}
\item \textbf{Attack Accuracy (AA)}: given an indicator function \(I_{xy}\):
\begin{equation}
    I_{xy} = \mathrm{1}\left[\forall (\tilde{x}, \tilde{y}) \in A(x, y): L(\tilde{x}) = \tilde{y}\right],
\end{equation}
\(L\)'s accuracy on the adversarial examples:
\begin{equation}
    \text{AA}(L) = \frac{\sum_{(x, y)\in P} I_{xy}}{|P|}.
\end{equation}
\item \textbf{Attack Success Rate (ASR)}: relative decrease in accuracy due to adversarial modifications, 
\begin{equation}
\text{ASR}(L) = \frac{\text{OA}(L) - \text{AA}(L)}{\text{OA}(L)}.
\end{equation}
\end{itemize}

\paragraph{Baseline}
As a baseline, we evaluate a common rephrasing approach focusing on modifying words within MWPs. \citet{zhou2023mathattack} freeze the logical entities and iteratively swap each unfrozen token with similar tokens that cause LLMs to fail. They then validate the adversarial example with human manual checking. \citet{zhou2023mathattack} created the RobustMath dataset, which includes 214 original problems from a combination of GSM8K and MultiArith with their 300 rephrased versions. We report the results in Appendix \ref{appendix:robustMath}, observing ASR for 4 out of the 7 models; however, some models even show improved accuracy given these adversarial examples, suggesting that such attacks may not generalize well across different LLMs.

\subsection{Our Attacks}
\paragraph{Problem Variation Generation}
We set the number of attempts to generate problem variants to be 30,000 and present the average number of generation per problem for each $M$ in Appendix \ref{appendix:num_generation}. While on average each problem generates over a thousand variants, it is worth noting that the actual number of possible generations varies based on the original values and the number of variable nodes in a problem.\footnote{For example, a problem with only one digit value and one variable node has a potential maximum of 9 possible modifications under $M3$.} Given this reason and cost considerations, in this work we randomly select 100 random problems from MultiArith and GSM8K and generate 100 variants for each selected problem in each $M$. Despite the seemingly small number of selected problems, the total data points are up to 60,000.\footnote{3 generation methods $\times$ 200 selected problems $\times$ 100 variants = 60,000 data points.}

\begin{table}[ht!]
\centering
\scriptsize
\begin{tabular}{lr}
\toprule
\multicolumn{1}{c}{\textbf{Models}} & \multicolumn{1}{c}{\textbf{\begin{tabular}[c]{@{}c@{}}Avg. (\%) $\downarrow$ \end{tabular}}} \\
\midrule
CodeLlama 34B & 91.8 \\
Llama 2 13B & 91.6 \\
Llama 3 8B &  78.3 \\
Mistral 7B & 70.5 \\
Vicuna 13B & 50.3 \\
WizardMath 13B & 21.0 \\
MetaMath 7B & 15.2 \\
MetaMath 70B & 7.9 \\
GPT-3.5 & 6.9 \\
\bottomrule
\end{tabular}
\caption{\textbf{Incorrect Variants:} The average percentage of incorrect variants per problem for each model in $M3$. Appendix \ref{appendix:distribution} shows detailed count distributions.}
\label{table:incorrect_variants}
\vspace{-2em}
\end{table}

\begin{table*}[ht!]
\centering
\scriptsize
\begin{tabular}{lrrr|rrr|r}
\toprule
& \multicolumn{3}{c|}{\textbf{RobustMath}} & \multicolumn{4}{c}{\textbf{Ours ($M3$)}} \\
\textbf{Model} & \textbf{OA} & \textbf{AA} & \textbf{ASR} & \textbf{OA} & \textbf{AA} & \textbf{ASR} & \textbf{$\Delta$ ASR} \\
\midrule
Mistral 7B & 10.3 & 18.7 & 0.0 & 33.0 & 0.0 & 100.0 & +100.0 \\
MetaMath 7B & 91.1 & 79.3 & 13.0 & 97.5 & 51.0 & 48.5 & +35.5 \\
Llama 3 8B & 22.0 & 30.0 & 0.0 & 19.0 & 0.0 & 100.0 & +100.0 \\
Llama 2 13B & 2.3 & 8.3 & 0.0 & 11.0 & 0.0 & 100.0 & +100.0 \\
WizardMath 13B & 71.0 & 70.3 & 1.0 & 89.0 & 15.5 & 82.6 & +81.6 \\
Vicuna 13B & 46.3 & 51.7 & 0.0 & 68.0 & 2.0 & 97.5 & +97.5 \\
CodeLlama 34B & 31.3 & 10.3 & 67.1 & 8.5 & 0.0 & 100.0 & +32.9 \\
MetaMath 70B & 93.0 & 82.7 & 11.1 & 98.5 & 68.0 & 31.1 & +20.0 \\
GPT-3.5 & 91.1 & 75.7 & 16.9 & 94.0 & 63.0 & 33.3 & +16.4 \\
\midrule
Average & 51.0 & 47.4 & 12.1 & 57.6 & 22.2 & 77.0 & \textbf{+62.0} \\
\bottomrule
\end{tabular}
\caption{\textbf{Comparison Result:} We calculate the average of each metric from both datasets and compare the result with rephrasing attack, RobustMath. Our method significantly outperforms the baseline in every model, with an average improvement of 62 ASR points.}
\label{table:comparison_result}
\vspace{-1em}
\end{table*}

\paragraph{Main Attacks}\label{sec:model_performance}
We report the model performance from all $M$s in Table \ref{table:main_results}.\footnote{We do not run the full dataset against GPT-4 due to cost considerations. Instead, we propose a cost-effective approach to query expensive models in \textit{Efficient Attack} section.
} Ranging from the most lenient method, $M1$, where the numbers could be fairly wild, to the most stringent one, $M3$, we observe a consistent performance drop across all models, even strong ones (e.g., math-tuned, larger size, or API-based). For the most original-problem-like generation, $M3$, weak models (e.g., smaller size or general-purpose) fail to generate \textit{any} correct answers. Table \ref{table:incorrect_variants} shows the average percentage of incorrect variants in each model, which is highly correlated with the reported ASR, with GPT-3.5 and MetaMath being the most robust and CodeLLama and LLama 2 the least.
Furthermore, comparing $M3$ with rephrasing attacks in the Table \ref{table:comparison_result}, our method significantly outperforms the baseline, resulting in a 62 ASR point improvement on average. 
We also analyze how ASR changes given different numbers of attacks in Appendix \ref{appendix:num_sample_graph}. Interestingly, for several models, just 10 attacks are enough to significantly degrade performance.

\begin{table}[ht!]
\scriptsize
\centering
\begin{tabular}{lrrrrr}
\toprule
\textbf{Model} & \textbf{Ct.}  & $\textbf{M3}$ \textit{{(\%)}}& $\textbf{M2}$ \textit{{(\%)}} & $\textbf{M1}$ \textit{{(\%)}} \\
\midrule
Mistral 7B& 1 & 100.0 & 100.0 & 100.0\\
MetaMath 7B & 2& 70.0 & 90.0 & 100.0 \\
Llama 3 8B & 3& 67.0 & 87.0 & 100.0 \\
Llama 2 13B & 4& 67.0 & 87.0 & 100.0 \\
WizardMath 13B & 5& 49.0 & 80.0 & 100.0 \\
Vicuna 13B & 6& 44.0 & 77.0 & 100.0 \\
CodeLlama 34B & 7& 19.0 & 65.0 & 99.0 \\
MetaMath 70B & 8& 9.0 & 49.0 & 87.0 \\
GPT-3.5 & 9& 9.0 & 46.0 & 83.0 \\
\bottomrule
\end{tabular}
\caption{\textbf{Universal Attack:} Universal attacks are shared among a number of models. Increasing the count of models being considered decreases the universal attacks, with $M3$ consistently showing the lowest percentages.}
\label{table:universal_attack}
\vspace{-1em}
\end{table}

\paragraph{Universal Attacks}
Following the previous adversarial attack literature, we investigate whether there are adversarial examples exist in all LLMs, also known as universal attacks \citep{zou2023universal,moosavi2017universal}. We count the number of adversarial examples from each model and calculated the percentage of common ones among them. We report the results on Table \ref{table:universal_attack} and observe a clear pattern of decreasing universal attacks with an increased number of models. This suggests that while all models are vulnerable to some degree of universal attacks, the percentage of such attacks decreases as more and diverse models are considered. We also observe that universal attacks varies significantly across different $M$s, with $M3$ having the lowest number of universal attacks.

\begin{table}[ht!]
\scriptsize
\centering
\begin{tabular}{lrrrr}
\toprule
\textbf{Models} & \textbf{Req Call} & \textbf{Cost (\$)} & \textbf{ASR} & \textbf{Req $\Delta$ } (\%) \\
\midrule
MetaMath 7B & 1,389 & 10.7& 8.0 & 72.2 \\
WizardMath 13B & 1,745 & 8.8& 6.0 & 65.1 \\
MetaMath 70B & 672 & 4.3& 8.0 & 86.6 \\
GPT-3.5  & 456 &2.8  & \textbf{10.0}& \textbf{90.9} \\
\midrule
Target: GPT-4 & 5,000 & 29.3 & 10 & - \\
\bottomrule
\end{tabular}
\caption{\textbf{Efficient Attack:} A targeted approach to attack high-cost models. We leverage adversarial examples from cheaper models to attack GPT-4, achieving the same ASR while reducing up to 90\% request calls.}
\label{table:efficient_attack}
\vspace{-1em}
\end{table}

\paragraph{Efficient Attacks}\label{sec:efficient_attack} 
Scaling up the number of attacks leads to a consistent performance drop for all models. However, this may not be feasible for API-based models like GPT-4 due to request limits and high costs. To address this, we propose an efficient attack method for API-based models by using adversarial examples in a \textit{targeted} manner. The idea is simple: we attack model $A$ (target model, GPT-4 in our case), with adversarial examples from a cheaper model $B$ (e.g., open-source, lower API cost). We select 50 problems along with their 100 variations and run them against GPT-4, comparing the results with attacking GPT-4 using 50 adversarial examples from a cheaper model. Note that for some models, there are no correct responses, even with $M3$. Therefore, we only compare $M3$ with the models that have correct answers. We compare the results in Table \ref{table:efficient_attack} and observe a significant request reduction while achieving similar ASR.

\section{Analysis and Discussion} 
In this section, we conduct analysis to better understand our generation methods and the mathematical capabilities of LLMs.

\begin{figure}[ht]
\small
\centering
\begin{minipage}{0.48\textwidth}
\centering
\includegraphics[width=\textwidth]{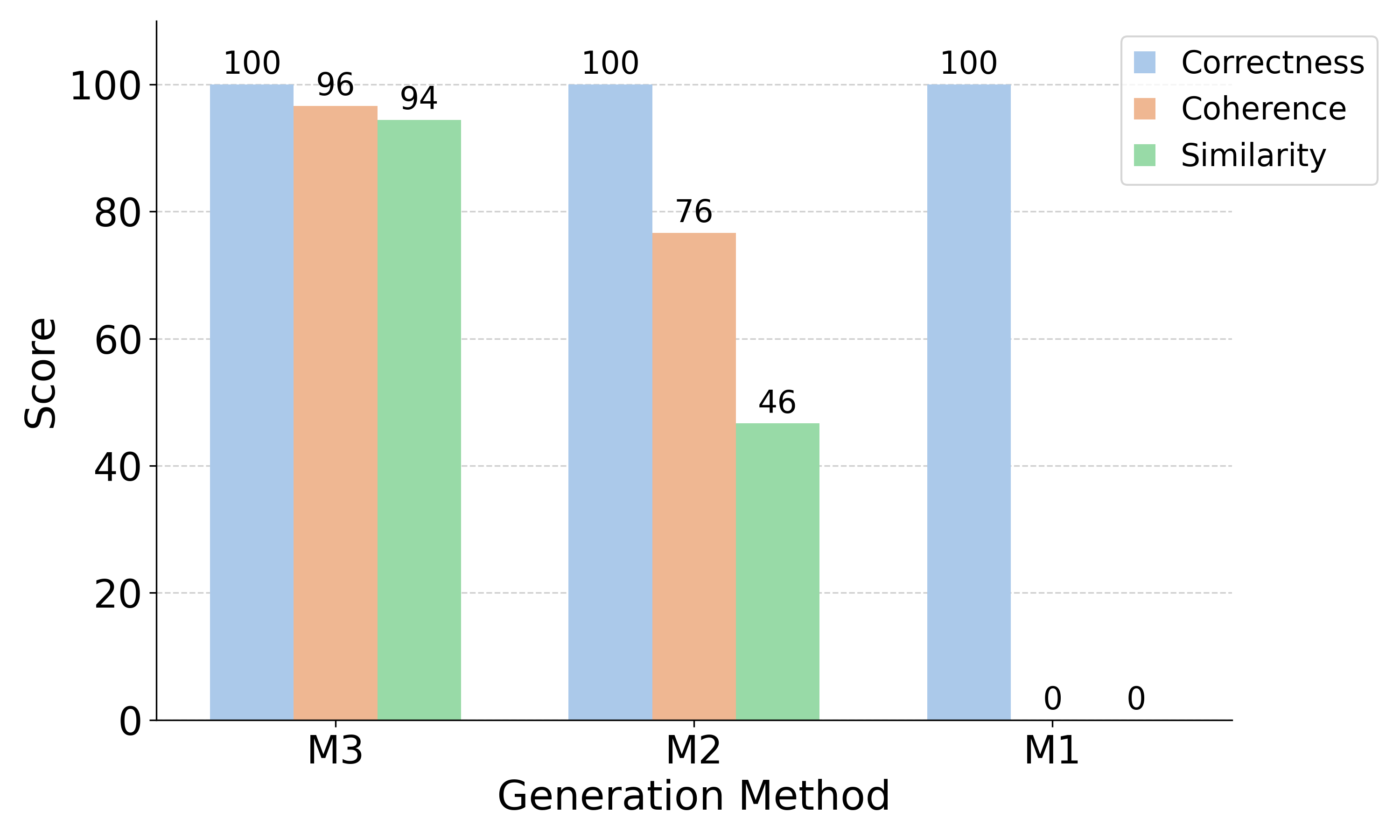}
\end{minipage}
\caption{\textbf{Human Evaluation:} The average score from three annotators. $M3$ achieves the highest scores across all metrics, indicating our best generation method correctly generates contextually coherent problems that preserve original difficulty.}
\label{fig:human_eval}
\vspace{-1em}
\end{figure}

\paragraph{Validation Through Human Evaluation} We conduct a human evaluation to ensure the validity of our generated problems. We randomly select 30 problems that GPT-3.5 failed on for GSM8K and select 2 adversarial examples from each problem. Three evaluators are asked to assess (i) correctness: whether the answer to the modified problem is correct; (ii) coherence: whether the problem is contextually coherent; and (iii) similarity: whether the newly generated problem's difficulty level matches the original. Evaluators make binary decisions to determine if the criteria are met, and Figure \ref{fig:human_eval} shows the average scores. Overall, all evaluators agree on correctness. $M3$ scores highly across \textit{all} metrics, indicating it preserves the original problem's coherence and difficulty. Coherence and similarity scores for $M2$ and $M3$ are relatively low, suggesting their perturbations might not be useful in an educational context.

\begin{figure}[ht]
\small
\centering
\begin{minipage}{0.48\textwidth}
\centering
\includegraphics[width=\textwidth]{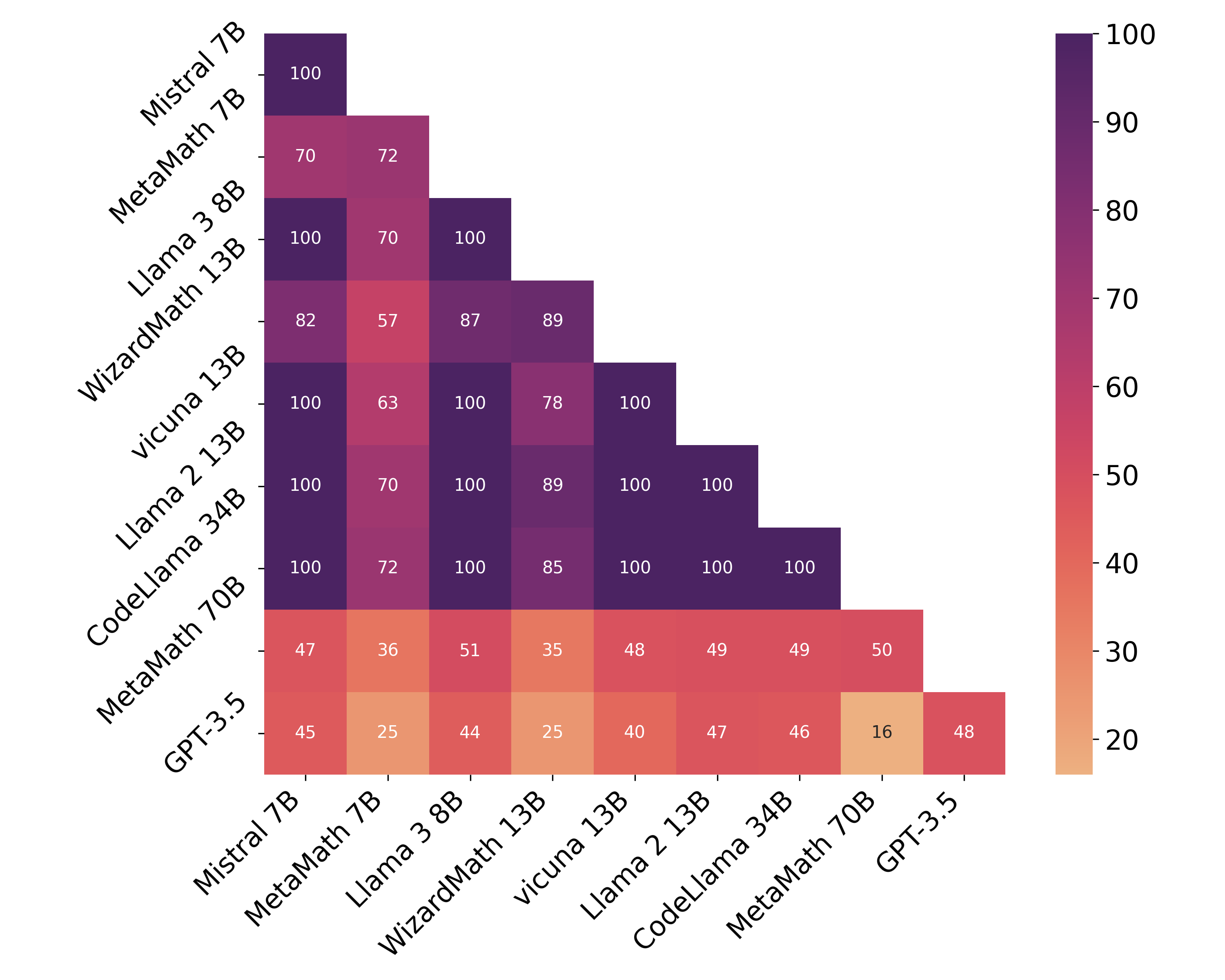}
\end{minipage}
\caption{\textbf{Transferability:} We present the adversarial example transferability (\%) among all models by comparing each model against all other models. Compared to the math-tuned and production models, the weaker models such as LLama2 13B exhibit significant vulnerability and a strong correlation among them.}
\label{fig:transferability}
\vspace{-1em}
\end{figure}

\paragraph{Transferability}
Following the previous work \cite{zou2023universal, zhou2023mathattack}, we investigate the transferability of adversarial examples among the models. Specifically, We attack model $A$ with the adversarial examples from model $B$ and calculate the number of common adversarial examples between these two models. 
We present $M3$ result in the Figure \ref{fig:transferability}. We observe that weaker models exhibit a high percentage of transferability, indicating a strong vulnerability correlation among them. On the other hand, strong models such as MetaMath 70B and GPT-3.5 tend to have a lower transferability rate. While it might suggest a form of resistance to adversarial examples that affect other models, it is not a measure of robustness as it could be that those models fail on entirely different examples.
This finding suggests that certain LLMs might struggle with particular numbers or patterns. To further understand these phenomena, we conduct regression analysis as shown below.

\paragraph{Regression Feature Analysis} \label{sec:failure_analysis}
To gain deeper insights into the limitations of LLMs on MWPs, we conduct a regression analysis investigating the relationships between various features of the problems and the correctness of the models' predictions. We construct a set of 51 input features from 20,000 $M3$ generated problems, including features such as operation counts, answer value ranges, and node counts in the problem's AST. By examining the coefficients of these features in predicting model correctness (see Table \ref{table:coefs_model} and Figure \ref{fig:accuracy} in the Appendix), we uncover interesting patterns that shed light on the limitations of LLMs:

\begin{itemize} [leftmargin=*,noitemsep,nolistsep]
\setlength{\itemsep}{1pt}
\setlength{\parskip}{1pt}
\setlength{\parsep}{1pt}

\item \textbf{Varying Vulnerabilities Across Models} We find that the most positively and negatively correlated features vary considerably across models, suggesting that our adversarial examples exploit distinct vulnerabilities. For instance, models exhibit divergent performance on problems with different answer value ranges. Mistral 7B and WizardMath 13B perform relatively well on problems with smaller answer values (e.g., in the range of [2, 8) and [8, 32)), while MetaMath 7B and Vicuna 13B show better performance on problems with answer values in the range of [32, 128). This observation hints that different models may have learned to specialize in problems with specific ranges, possibly due to variations in the distributions of their pre-training data.

\item \textbf{Complexity and Operation Types} We observe that problems involving division and a higher number of operations tend to be more challenging for most LLMs (see Figure \ref{fig:accuracy} in the Appendix for visualizations). For instance, problems requiring multiple division operations or a combination of different operations (e.g., addition, subtraction, multiplication, and division) are more likely to result in incorrect predictions. This aligns with the intuition that complex problems requiring more mathematical operations are generally more difficult for LLMs to solve \citep{imani2023mathprompter}. 

\item \textbf{Tokenization Choices and Numerical Reasoning:} Recent research has also highlighted the impact of tokenization choices on LLMs' numerical reasoning capabilities \citep{singh2024tokenization}. Our experiments reveal similar patterns, with LLama-based models tokenizing each digit individually, while GPT-3.5 and GPT-4 encode every three digits into a single token. This difference in tokenization may contribute to the observed consistent performance of GPT-3.5 to the number of tokens in the answer compared to the LLama-based models.
\end{itemize}

\paragraph{Why LLMs suffer From Such Simple Attacks?}
While LLMs can be trained on vast amounts of math data, they may not have an inherent grasp of the underlying mathematical concepts and reasoning steps \citep{saxton2019analysing,lample2019deep}. Instead, 
they likely rely on pattern and statistical associations learned from the training data to make predictions \citep{10.1145/3442188.3445922}, which is also know as memorization \citep{carlini2022quantifying}. Modifying the numbers in a problem may disrupt these learned patterns, causing the models to make errors. This is evidenced by the significant performance drops observed in models like WizardMath (89\% to 20\%) and Vicuna (76\% to 4\%) when presented with adversarial examples (Table \ref{table:main_results}). Furthermore, test data contamination could also play a role - if the training corpus contains the test data, the models may be able to answer the original questions correctly but struggle to generalize to even subtle variations \citep{balloccu2024leak}.


\paragraph{Generated Python Code and ASTs Analysis} \label{sec:python_code_analysis}
To ensure that the generated Python code is valid and can be correctly converted into ASTs, we manually examined 200 Python code and ASTs pairs from GSM8K and MultiArith, respectively. We identify the four types of errors:
\begin{itemize} [leftmargin=*,noitemsep,nolistsep]
\setlength{\itemsep}{1pt}
\setlength{\parskip}{1pt}
\setlength{\parsep}{1pt}
\item \textbf{Incorrect answers or code:} The generated code produces incorrect answers or contains logical errors that deviate from the problem statement.
\item \textbf{Use of unsupported constructs:} The code includes loops, conditional and comparison statements, or user-defined functions, which are not supported by our AST conversion process.
\item \textbf{Complex expressions:} The code contains expressions that are difficult to convert into ASTs, such as \verb|x = y // z + (y % z > 0)|.
\item \textbf{Number misalignment:} The same number appears multiple times in the code or the problem, leading to inconsistencies between the generated code and the problem statement.
\end{itemize}
Encouragingly, we find that 96.5\% and 92.5\% of the generated ASTs for GSM8K and MultiArith, respectively, are valid and free from these errors. We discuss this process in more details in the Appendix \ref{sec:app_code_ana}. The high percentage of valid code and ASTs pair indicates that our code generation approach is indeed reliable.

\section{Conclusion}
We introduce a novel method to generate adversarial MWPs using ASTs. Our approach effectively challenges the mathematical problem-solving abilities of LLMs while maintaining the original difficulty and coherence of the problems. The generated adversarial examples significantly degrade the performance of both open- and closed-source LLMs, surpassing previous attacks by an average of 62\% ASR. We validate our adversarial examples through human evaluation and investigate universal attacks and transferability, proposing a cost-effective method to attack high-cost API-based models with up to a 90\% reduction in requests. Automatic regression analyses reveal distinct weaknesses of different models when solving MWPs. Our work contributes to the development of fair and robust educational tools, ensuring ethically sound evaluations and promoting the responsible use of LLMs in education.

\section{Limitations}
This work has not empirically validated the correlation between the complexity of generated problems and the actual difficulty perceived by human students. Further investigation is needed to establish this correlation.

As a proof of concept, our method demonstrates the feasibility of generating meaningful and useful adversarial examples to LLMs, at least at the difficulty levels represented by the GSM8K and MultiArith datasets. However, it may not generalize well to math problems with significantly different types or difficulty levels. For example, very simple arithmetic problems (e.g., 3 + 2 = ?) might never have a successful attack on LLMs specifically trained for math, while advanced problems requiring proofs or concepts without numeric manipulation (e.g., prove newton's binomial theorem) may not be suitable for our code generation procedure. As LLMs continue to improve, it is possible that they may become more robust, which is a challenge faced by \textit{all} adversarial attack methods in general. We plan to address these limitations in future work by exploring a wider range of math problems and investigating other attack vectors as LLMs evolve.

\section{Ethics Statement}
While our primary intention is to address concerns about academic dishonesty, we acknowledge that our strategy might also lead to potential exacerbating educational inequality. By designing math problems specifically to be unsolvable by LLMs, the educators without access to resources, training, or the necessary tools could be placed at a disadvantage. This approach risks amplifying the disparities between institutions or individuals with differing levels of technological access. 

\section{Acknowledgements}
We are very grateful to Sandy Yeung and Jack Goffinet for the helpful initial project discussion. Thank you to the anonymous reviewers for their feedback. The research reported here was supported by the Learning Engineering Virtual Institute funded by leading education philanthropists and organizations through Grant G-23-2137070, to the University of Florida and its partner institutions. The opinions expressed are those of the authors and do not represent the views of the University of Florida, the partner institutions, or those of the philanthropists and organizations.

\bibliography{custom}

\clearpage
\newpage
\appendix
\onecolumn

\section{Prompt Templates}
\subsection{Code Generation Prompt}\label{appendix:code_gen_prompt}

\begin{tcolorbox}[]\small
Write a Python script that contains a step-by-step solution for a given math problem and its answer. Start the script by defining variables corresponding to the quantities mentioned in the problem. Print the final answer at the end of the script. Do not use if-else statements, floor divisions, and loops to solve the problem.
\\ \\
Given problem: \{problem\} 
\\ \\
Answer: \{answer\}
\\ \\
Python script:

\end{tcolorbox}

\subsection{Zero-shot CoT}\label{appendix:main_prompt}
\begin{tcolorbox}[]\small
Solve a math problem. The solution ends with ``the answer is (a number)" like ``the answer is 1".
\\ \\
Question: \{problem\} 
\\ \\
Answer: Let's think step by step.

\end{tcolorbox}

\section{A Rephrasing Attack Example}\label{appendix:rephrasing_example}
\begin{figure}[h!]
    \centering  
    \includegraphics[width=12cm]{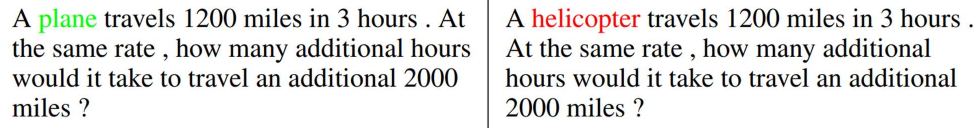}
  \caption{A rephrasing attack example from \citet{zhou2023mathattack}. The left side shows the original problem, and the right side is the rephrased version. Although planes and helicopters are conceptually similar, the maximum flying distance for a helicopter is usually between 300 to 400 miles. The rephrasing introduces a subtle and incorrect factual error into the problem, which could be hard to detect at times.}
  \vspace{-1em}
\end{figure}
\section{AST Nodes}\label{appendix:node_detail}
With the two main types of nodes, we further distinguish them into following nodes: 
\begin{itemize}[leftmargin=*,noitemsep,nolistsep]
  \setlength{\itemsep}{1pt}
  \setlength{\parskip}{1pt}
  \setlength{\parsep}{1pt}
    \item \textbf{Binary operation node}: A node consists of two operands (nodes) and one main operation. 
    \item \textbf{Unary operation node}: A node consists of only one operand and one main operation.
    \item \textbf{Variable node}: Represents the corresponding variable and its value from the problem. The set of variable nodes will be represented by $V=\{v_1,\dots,v_n\}\subset S$, where $S$ represents the full tree. 
    \item \textbf{Constant node}: A special type of variable node, where the number appearing in the code does not correspond to any number in the problem. For example, a problem mentioning ``one year'' implies a number of ``365'' days in the code. 
\end{itemize}
We associate the number in the original question with the variable node so that a change in the variable node will also change the number in the question. If a number appears multiple times in the code and the problem, then we associate them based on the specific procedures described in Section \ref{sec:num_mis}.
\section{Generation Details}\label{appendix:gen_detail}
In this section, we describe the generation detail for new numbers in each generation method. Each variable node in an AST generates a new number $val(v') = p'$ by following Algorithm \ref{algo:new_num_gen}, which takes the AST $S$, a maximum number of attempts $N$, a maximum number of scientific digits $c$, the set of Boolean constraints $C$, and the desired generation $M$ method as inputs. It produces $m$ sets of new problems $W'=\{V'_1,\dots,V'_m\}$ that satisfy the given node constraints defined in \sref{sec:node_constraints}.A newly generated adversarial example is valid if all nodes have their given constraints satisfied. Unless otherwise stated, we use $N$=30,000 and $c$=6 for all experiments.
\begin{algorithm}[ht]
\small
\SetAlgoNoLine 
\caption{Number Generation Based on Method}
\label{algo:new_gen_con}

\DontPrintSemicolon
\KwIn{$v_i, M, c$}
\KwOut{$v'_i$}
Let $a_i \times 10^b_i\approx h = val(v_i)$, where $a_i$ is an integer that is not divisible by 10 and has at most $c$ scientific digits.\;
\uIf{$M = \textbf{\upshape M1}$}{
    \eIf{$v_i$ is a divisor for a division node}{
        $a'_i \sim \text{Uniform}(1, 10 ^ {c/2})$\;
    }{
        $a'_i \sim \text{Uniform}(1, 10 ^ {c})$\;
    }
}\uElseIf{$M = \textbf{\upshape M2}$}{
    Let $d$ be the number of digits $val(v_i)$ has.\;
    \uIf{$d = 1$}{
        $a'_i \sim \text{Uniform}(1, 99)$\;
    }\uElse{
        $a'_i \sim \text{Uniform}(10^{d-1}, 10^{d} - 1)$\;
    }
    $b\leftarrow0$ if $b>0$ else $b$\;
}\uElseIf{$M = \textbf{\upshape M3}$}{
    \eIf{$1 \leq a_i \leq 9$}{
        $a'_i \sim \text{Uniform}(1,9)$\;
    }{
        $a'_i \sim \text{Pois}(a_i)$\;
    }
}
$val(v'_i) \leftarrow a'_i\times 10^b$\;
\Return{$v'_i$}
\end{algorithm}

\begin{algorithm}[ht]
\small
\SetAlgoNoLine 
\caption{Problems Generation}
\label{algo:new_num_gen}

\DontPrintSemicolon
\KwIn{$S=\{s_1,\dots,s_u\}$, $V=\{v_1,\dots,v_n\}\subset S$, $N$, $c$, $C$, $M$}
\KwOut{$W'=\{V'_1,\dots,V'_m\}$, where $V'_j=\{v'_1,\dots,v'_n\}$}

$attempts \leftarrow 0, j \leftarrow 1$\;
$W' = \emptyset $\;
\While{$attempts < N$}{
    $V'_j = \emptyset $\;
    \For {$v_i\in V$} {
        Obtain $v'_i$ from Algorithm \ref{algo:new_gen_con} with input $v_i$, $M$, and $c$\;
        $V'_j\leftarrow V'_j\cup\{v'_i\}$\;
    }
    $attempts\leftarrow attempts + 1$\;
    $S' \leftarrow S$\;
    Apply the new sequence of values from the variable nodes $V'_j$ to $S'$\;
    $accept\leftarrow True$\;
    \For {each node $s'_k$ in $S'$} {
        \If{$\exists C_i\in C, C_i(s_k) \centernot\implies C_i(s'_k)$}{
            $accept\leftarrow False$\;
        }
    }
    \If{accept}{
        $W'\leftarrow W'\cup \{V'_j\}$\;
        $j\leftarrow j+1$\;
    }
}
\Return{$W'$}

\end{algorithm}

\begin{table}[ht!]
   \footnotesize 
    \centering
  \begin{tabular}{|l|l|l|}
    \hline
    \textbf{Method} & \textbf{Question} & \textbf{Answer}  \\
\hline
Original & \begin{tabular}[c]{@{}l@{}}Mary does her grocery shopping on Saturday. She \\ does her shopping only at a specific store where \\ she is allowed a credit of \$100, which must be paid \\ in full before her next shopping trip. That week she \\ spent the full credit limit and paid \$15 of it on \\ Tuesday and \$23 of it on Thursday. How much \\ credit will Mary need to pay before her next \\ shopping trip?\end{tabular}        & 62     \\ \hline
M3       & \begin{tabular}[c]{@{}l@{}}Mary does her grocery shopping on Saturday. She \\ does her shopping only at a specific store where she \\ is allowed a credit of \$80, which must be paid in full \\ before her next shopping trip. That week she spent \\ the full credit limit and paid \$12 of it on Tuesday and \\ \$19 of it on Thursday. How much credit will Mary \\ need to pay before her next shopping trip?\end{tabular}            & 49     \\ \hline
M2       & \begin{tabular}[c]{@{}l@{}}Mary does her grocery shopping on Saturday. She \\ does her shopping only at a specific store where she \\ is allowed a credit of \$432, which must be paid in \\ full before her next shopping trip. That week she \\ spent the full credit limit and paid \$91 of it on \\ Tuesday and \$76 of it on Thursday. How much \\ credit will Mary need to pay before her next \\ shopping trip?\end{tabular}        & 265    \\ \hline
M1       & \begin{tabular}[c]{@{}l@{}}Mary does her grocery shopping on Saturday. She \\ does her shopping only at a specific store where she \\ is allowed a credit of \$56347, which must be paid in \\ full before her next shopping trip. That week she \\ spent the full credit limit and paid \$54731 of it on \\ Tuesday and \$1566 of it on Thursday. How much \\ credit will Mary need to pay before her next \\ shopping trip?\end{tabular} & 50     \\ \hline
Original & \begin{tabular}[c]{@{}l@{}}A bird watcher records the number of birds he sees each \\ day. One Monday he sees 70 birds. On Tuesday he sees \\ half as many birds as he did on Monday. On Wednesday \\ he sees 8 more birds than he did on Tuesday. How many \\ total birds did the bird watcher see from Monday to\\ Wednesday?\end{tabular}         & 148    \\ \hline
M3       & \begin{tabular}[c]{@{}l@{}}A bird watcher records the number of birds he sees each \\ day. One Monday he sees 80 birds. On Tuesday he sees \\ half as many birds as he did on Monday. On Wednesday \\ he sees 2 more birds than he did on Tuesday. How many \\ total birds did the bird watcher see from Monday to \\ Wednesday?\end{tabular}        & 162    \\ \hline
M2       & \begin{tabular}[c]{@{}l@{}}A bird watcher records the number of birds he sees each \\ day. One Monday he sees 26 birds. On Tuesday he sees \\ half as many birds as he did on Monday. On Wednesday \\ he sees 39 more birds than he did on Tuesday. How many \\ total birds did the bird watcher see from Monday to \\ Wednesday?\end{tabular}       & 91     \\ \hline
M1       & \begin{tabular}[c]{@{}l@{}}A bird watcher records the number of birds he sees each \\ day. One Monday he sees 57010 birds. On Tuesday he sees \\ half as many birds as he did on Monday. On Wednesday he \\ sees 86391 more birds than he did on Tuesday. How many \\ total birds did the bird watcher see from Monday to \\ Wednesday?\end{tabular} & 200411 \\ \hline
  \end{tabular}
    \caption{Examples of questions generated by the three generation methods. }
    \label{appendix:problem_sample}

\end{table}

\clearpage
\newpage

\section{Rephrasing Attack Performance}  \label{appendix:robustMath}

\begin{table}[ht!]
  \small
    \centering
  \begin{tabular}{lrrr}
    \toprule
    \textbf{Model} & \textbf{OA} & \textbf{AA} & \textbf{ASR}  \\
    \midrule
    Mistral 7B &  10.3 &  18.7  & 0.0 \\
    MetaMath 7B &  91.1 &  79.3  & 13.0 \\
    Llama 3 8B & 22.0 & 30.0 & 0.0\\
    Llama-2 13B &  2.3 &  8.3  & 0.0 \\
    WizardMath 13B & 71.0  &  70.3  & 1.0 \\
    Vicuna 13B &  46.3 &  51.7  & 0.0 \\
    CodeLlama 34B & 31.3  &  10.3  & 67.1 \\
    MetaMath 70B &  93.0 & 82.7   & 11.1 \\
    GPT-3.5 &  91.1 &  75.7  & 16.9 \\
    \bottomrule
  \end{tabular}
    \caption{While rephrasing the problem results in lower accuracy for 4 out of 7 models, a few models show improved accuracy with the rephrased version. This suggests that rephrasing the MWPs may not generalize effectively across different LLMs.
    }
  \vspace{-1em}
\end{table}

\section{Incorrect Variant Count Distribution}\label{appendix:distribution}

\begin{figure}[h]
\small
    \centering  
    \includegraphics[width=\linewidth]{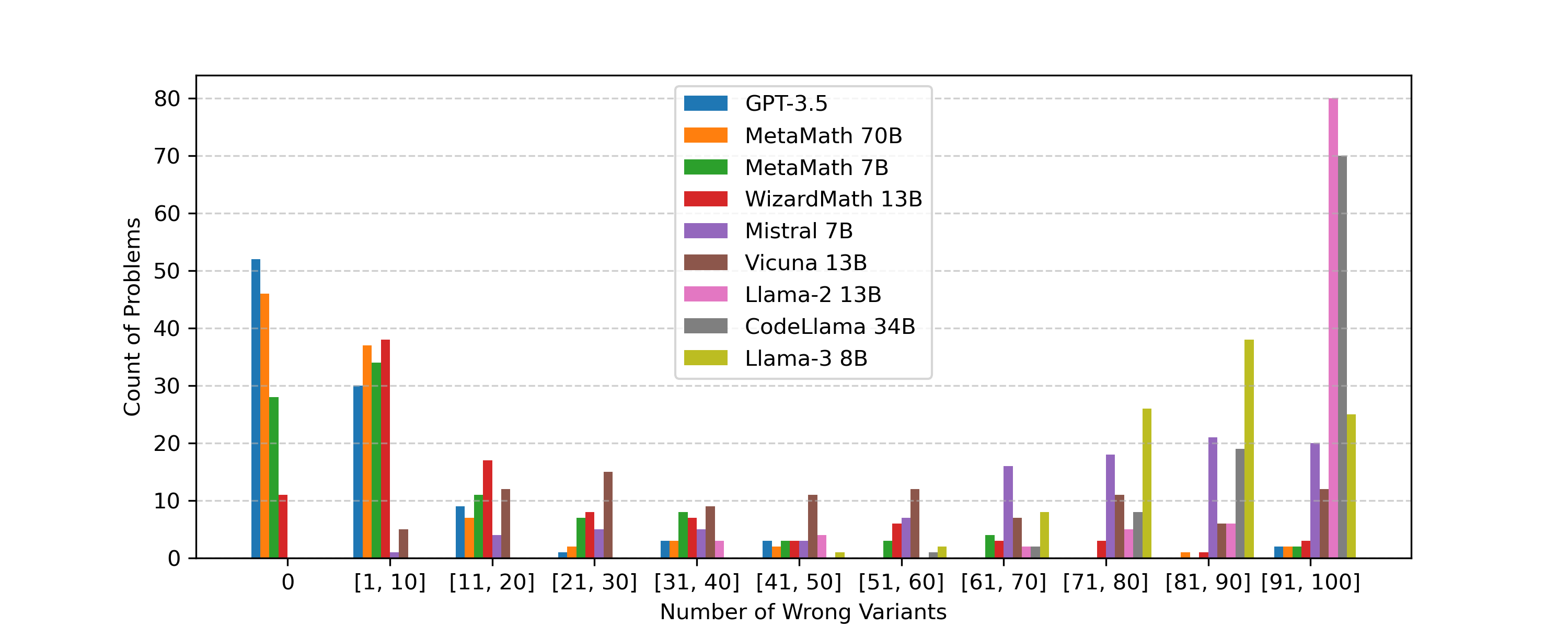}
    \caption{We present the distribution of incorrect adversarial examples in different buckets. For strong models (e.g., GPT-3.5-Turbo, MetaMath 70B), around half of the problems have zero incorrect adversarial examples, while for weaker models (e.g., Llama-2 13B, CodeLlama 34B), most problems have more than 90 incorrect adversarial examples.
    }
    \vspace{-1em}
\end{figure}
\section{Number of Generation}\label{appendix:num_generation}

\begin{figure}[h!]
\small
    \centering  
    \includegraphics[width=7.5cm]{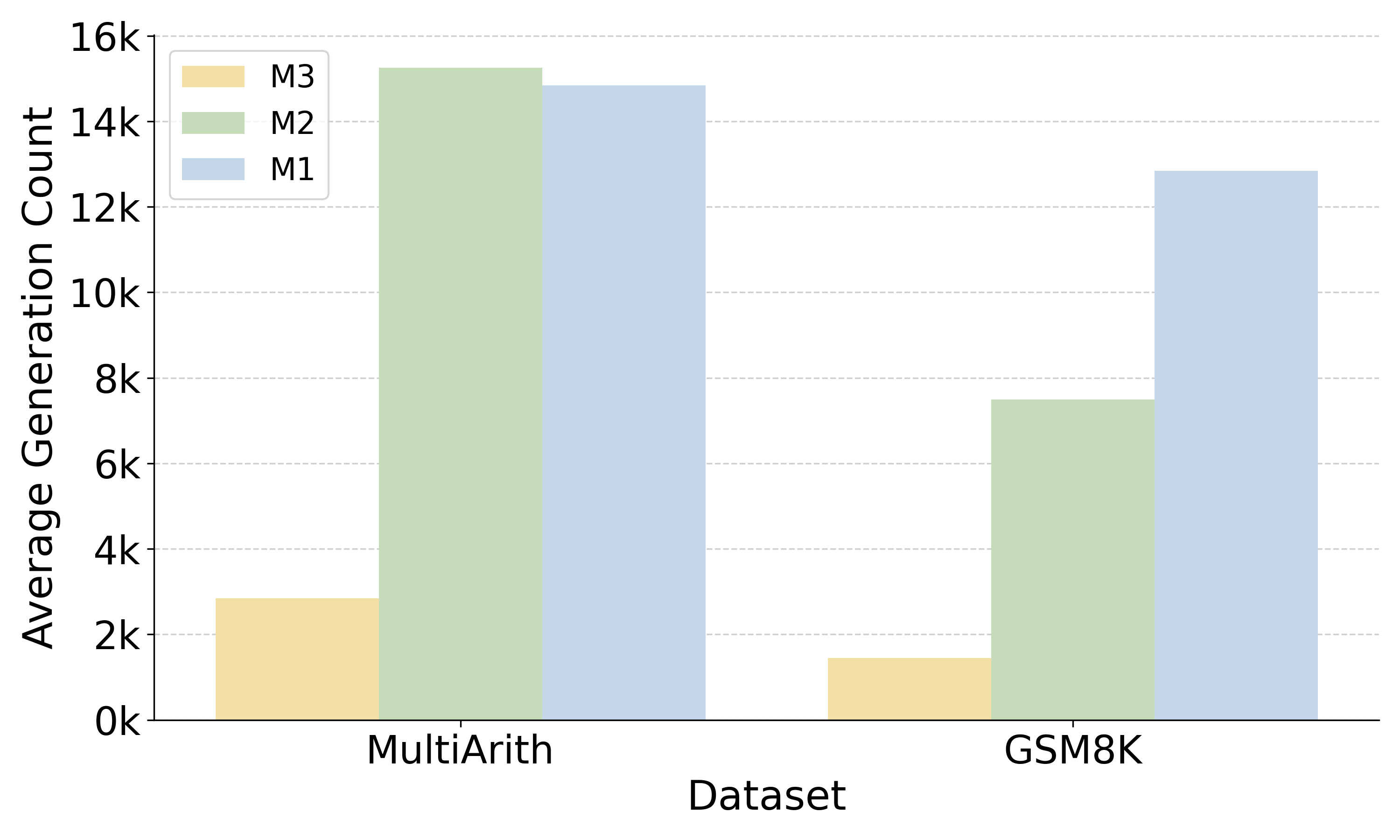}
    \caption{With 30,000 attempts for each given constraint, we calculate the average number of adversarial examples generated for each problem.}
    \label{fig:num_avg_count}
    \vspace{-1em}
\end{figure}

\section{Code Analysis} \label{sec:app_code_ana}

\subsection{Incorrect Answer from the Dataset} 
Given the execution steps in code are almost always deterministic, we are able to identify this error effectively. There are also instances where the dataset-provided answers might be wrong. An example of such an instance is the following question from the MultiArith dataset: "Paige's team won their dodgeball game and scored 41 points total. If Paige scored 11 of the points and everyone else scored 6 points each, how many players were on her team?", which the dataset provided 5 as the answer. However, the correct answer should be 6, as Paige is on the team as well. The code provided by GPT-4 counted Paige to the team. To automatically detect such instances, the value of the final answer node will be compared to the answer given by the dataset, and the question will be filtered out for further generations of problems if the two answers don't match. In total, we found only 1\% and 2.5\% generated codes have incorrect logic or answers in GSM8K and MultiArith, respectively.

\subsection{Incorrect Answer from GPT-4}\ 

Similarly, there are also instances where GPT-4 generated codes that provided the wrong answer to the given question. One such question is from the GSM8K dataset: "Jen decides to travel to 3 different countries.  He has to pay \$400 for the supplies he needs, in total.  The tickets for travel cost, in total, 50\% more than the supplies.  How much does travel cost?" The correct answer should be $400 + 400 \times 1.5 = 1000$, but the code generated by GPT-4 outputted \$600. Similar to the previous error, instances where the code outputs the wrong answers will be detected by comparing them to answers given by the dataset, and the instances will be discarded from generating math problems.

\subsection{Contains Control Flow Statements} 
Although specified in the prompt, GPT-4 might generate codes that include control flow statements (including "if", "for", and "while" statements). Codes that contain control flow statements are considered errors and discarded because the codes are usually only applicable to the original instance of the question. Furthermore, the code might not halt for some combinations of numbers, resulting in no answers associated with the generated instances.

\subsection{Number Misalignment} \label{sec:num_mis}
To ensure the correct generation of new questions, numbers in the original question need to be associated with numbers in the code. Given the sequence of numbers in the question $(q_1,\dots,q_n)$ and the sequence of numbers in the code $(c_1,\dots,c_m)$, the goal of grounding is to associate $q_i$ to the correct $c_j$ in the code. For distinct $q_i$'s, such pairing is trivial. However, there are instances where some numbers appear multiple times in the question or the code, where a strategy for grounding is needed. Let $C_q(q_i)$ be the count of $q_i$ in the question, and let $C_c(c_i)$ be the count of $c_i$ in the code. The strategy will be broken down into several scenarios:
\begin{enumerate}
    \item If there is a number $q_i$, such that $C_q(q_i) < C_c(q_i)$, then the question will be discarded for further problem generations. The reason to discard this kind of problem is that the relation between the numbers can be very convoluted. One number in the question might correspond to zero, one, or many numbers in the code, and finding the correct correspondence requires human evaluation. 
    \item There is a number $q_i$, such that $C_q(q_i)>1$ and $C_q(q_i) = C_c(q_i)$. Similar to the previous scenario, one number in the question can correspond to zero, one, or many numbers in the code. However, after some manual inspection, we find that most questions under this scenario have a one-to-one correspondence between the numbers in the questions and the codes. Therefore, we have devised a method to ensure that the right correspondence of numbers can be retrieved. Given that the number appears $C_q(q_i)$ times in the question and the code, we constructed $C_q(q_i)!$ pairs of one-to-one correspondences to the numbers from the syntax tree to the question. Then, random numbers are generated to combine the equivalences of correspondences. For example, given that the question is "Mary has 5 apples and 5 oranges, how many fruits does Mary have?". Although the two "5"s correspond to different representations in the question, due to the communicative nature of addition, the two "5"s are interchangeable in the syntax tree. Thus, the two pairs of correspondence for this question can be combined, and there is no ambiguity in the grounding. After the correspondences are combined, one valid sequence of numbers for the question is randomly generated for the code that ensures that all correspondences of the syntax tree output different answers. Then, this version of the question is asked to GPT-4 to obtain the answer to the question. The correspondence that has the same answer will be used, and if no correspondence has the correct answer, the question will be discarded. An example of this type of question is: "The pizzeria sells small pizzas for \$2 and large pizzas for \$8. They sold \$40 in pizzas. If they sold 8 small pizzas, how many large pizzas did they sell?". One valid and distinguishing sequence of numbers for the question is $(2, 3, 31, 5)$, which will result in different answers $5$ and $7$.
    \item There is a number $c_i$, such that $C_q(c_i)>C_c(c_i)$. Similar to the reasoning for the first case, these kinds of questions are eliminated for problem generations.
\end{enumerate}

\section{Number of Attacks}\label{appendix:num_sample_graph}

\begin{figure}[h!]

    \centering  
    \includegraphics[width=12cm]{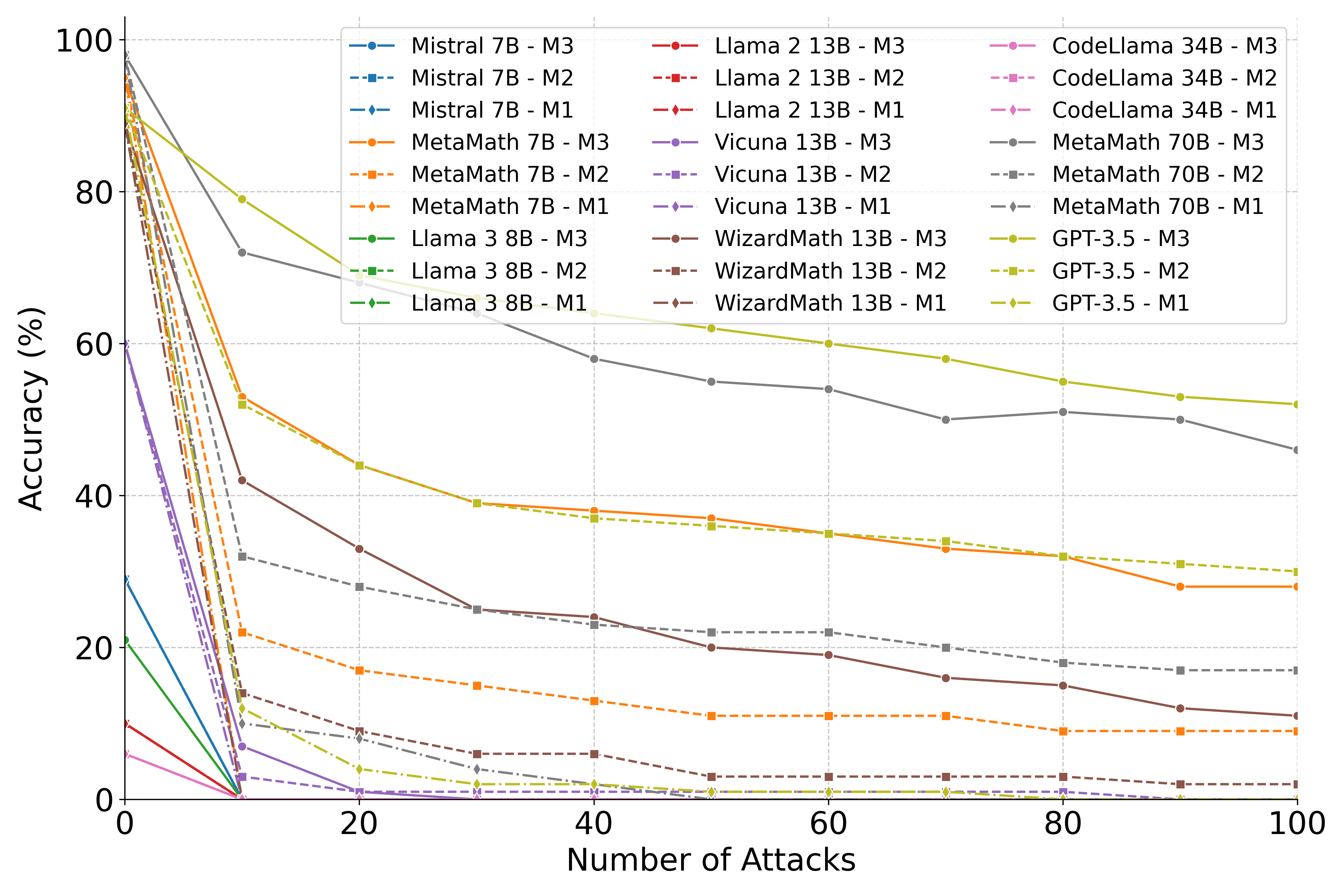}
    \caption{We report the model performance given various number of attacks. For several models, only 10 attacks are enough to degrade the performance.}
    \label{fig:num_sample_graph}
    \vspace{-1em}
\end{figure}

\clearpage
\newpage
\section{Feature Analysis}\label{appendix:feature_analysis}

\begin{table}[ht]
  \small
    \centering
  \begin{tabular}{lrrrr}
    \toprule
    \textbf{Features}& \textbf{MetaMath 7B} & \textbf{Vicuna 13b} & \textbf{CodeLlama 34b} & \textbf{GPT-3.5} \\
    \midrule
Addition Count      & -0.0032                   & 0.0080                   & -0.0146                   & 0.0098                      \\
Divide Count        & -0.0648                   & \textbf{-0.1137}         & 0.0142                    & \underline{-0.0804}         \\
Minus Count         & -0.0187                   & 0.0040                   & -0.0254                   & -0.0195                     \\
Multiply Count      & -0.0328                   & 0.0214                   & 0.0031                    & -0.0520                     \\
Constant Count      & 0.0923                    & 0.0782                   & 0.0027                    & 0.1559                      \\
Variable [8, 32)    & 0.0011                    & -0.0142                  & -0.0100                   & 0.0117                      \\
Answer [2, 8)       & 0.2215                    & 0.1377                   & \textbf{0.0632}           & -0.0736                     \\
Answer [8, 32)      & 0.2437                    & 0.1104                   & 0.0215                    & -0.0787                     \\
Answer [32, 128)    & \underline{0.2610}        & \underline{0.1670}       & 0.0183                    & -0.0508                     \\
Answer [128, 512)   & 0.2267                    & 0.0998                   & -0.0259                   & -0.0726                     \\
Answer [512, 2048)  & 0.0076                    & 0.0864                   & \underline{-0.0380}       & -0.0343                     \\
Answer [2048, 8192) & \textbf{-0.1987}          & -0.0775                  & \textbf{-0.0434}          & 0.0336                      \\
Convert to Int      & 0.2400                    & 0.1664                   & \underline{0.0470}        & \underline{0.2476}          \\
Operation Count     & \underline{-0.1196}       & \underline{-0.0804}      & -0.0227                   & \textbf{-0.1421}            \\
Variable Count      & 0.0722                    & 0.0254                   & 0.0074                    & 0.0939                      \\
Constant            & \textbf{0.2840}           & \textbf{0.1840}          & 0.0328                    & \textbf{0.3919}             \\
\midrule
\textbf{Features}   & \textbf{Llama 2 13b} & \textbf{MetaMath 70B} & \textbf{Mistral 7B} & \textbf{WizardMath 13B} \\
    \midrule
Addition Count      & -0.0163                 & -0.0287                    & 0.0102                   & -0.0118                      \\
Divide Count        & \textbf{0.0523}         & 0.0135                     & 0.0134                   & -0.0423                      \\
Minus Count         & -0.0152                 & -0.0271                    & 0.0097                   & -0.0137                      \\
Multiply Count      & -0.0248                 & \underline{-0.0473}        & \underline{-0.0422}      & -0.0369                      \\
Constant Count      & -0.0051                 & 0.1136                     & 0.0478                   & 0.0958                       \\
Variable [8, 32)    & \underline{-0.0349}     & 0.0286                     & -0.0383                  & 0.0266                       \\
Answer [2, 8)       & 0.0205                  & 0.0588                     & \underline{0.1006}       & \underline{0.1448}           \\
Answer [8, 32)      & \underline{0.0256}      & 0.0930                     & \textbf{0.1065}          & 0.1199                       \\
Answer [32, 128)    & -0.0011                 & 0.1186                     & 0.0479                   & 0.1234                       \\
Answer [128, 512)   & -0.0006                 & 0.1506                     & 0.0325                   & 0.0964                       \\
Answer [512, 2048)  & -0.0186                 & 0.0771                     & -0.0106                  & -0.0637                      \\
Answer [2048, 8192) & -0.0242                 & 0.0105                     & -0.0008                  & \textbf{-0.1885}             \\
Convert to Int      & \textbf{-0.0435}        & \underline{0.1771}         & \textbf{-0.0894}         & 0.0498                       \\
Operation Count     & -0.0040                 & \textbf{-0.0895}           & -0.0089                  & \underline{-0.1047}          \\
Variable Count      & 0.0217                  & 0.1068                     & 0.0238                   & 0.0795                       \\
Constant            & 0.0205                  & \textbf{0.3100}            & 0.0805                   & \textbf{0.2800}              \\
    \bottomrule
  \end{tabular}
    \caption{Coefficients of regression analysis on selected features for each model. Positive coefficients indicate that the model performs better on problems with the corresponding feature, while negative coefficients indicate the opposite. The most positive and negative correlation coefficients for each model are \textbf{bolded}, and the second most positive and negative correlation coefficients are \underline{underlined}. We observe that the most positively and negatively correlated features vary across different models, suggesting that our adversarial examples exploit different weaknesses of each model. For example, Mistral 7B and WizardMath 13B perform relatively well on problems with smaller answer values (e.g., in the range of [2, 8) and [8, 32)), while MetaMath 7B and Vicuna 13B show better performance on problems with answer values in the range of [32, 128). The observation matches the accuracy shown in Figure \ref{fig:accuracy}. Note that coefficients should not be compared across LLMs, only with other features within the same model. A more suited metric to compare between LLMs is accuracy, such as Table \ref{table:main_results} and Figure \ref{fig:num_sample_graph}.}
  \label{table:coefs_model}
  \vspace{-1em}
\end{table}

\begin{figure}[ht]
    \centering  
    \includegraphics[width=\linewidth]{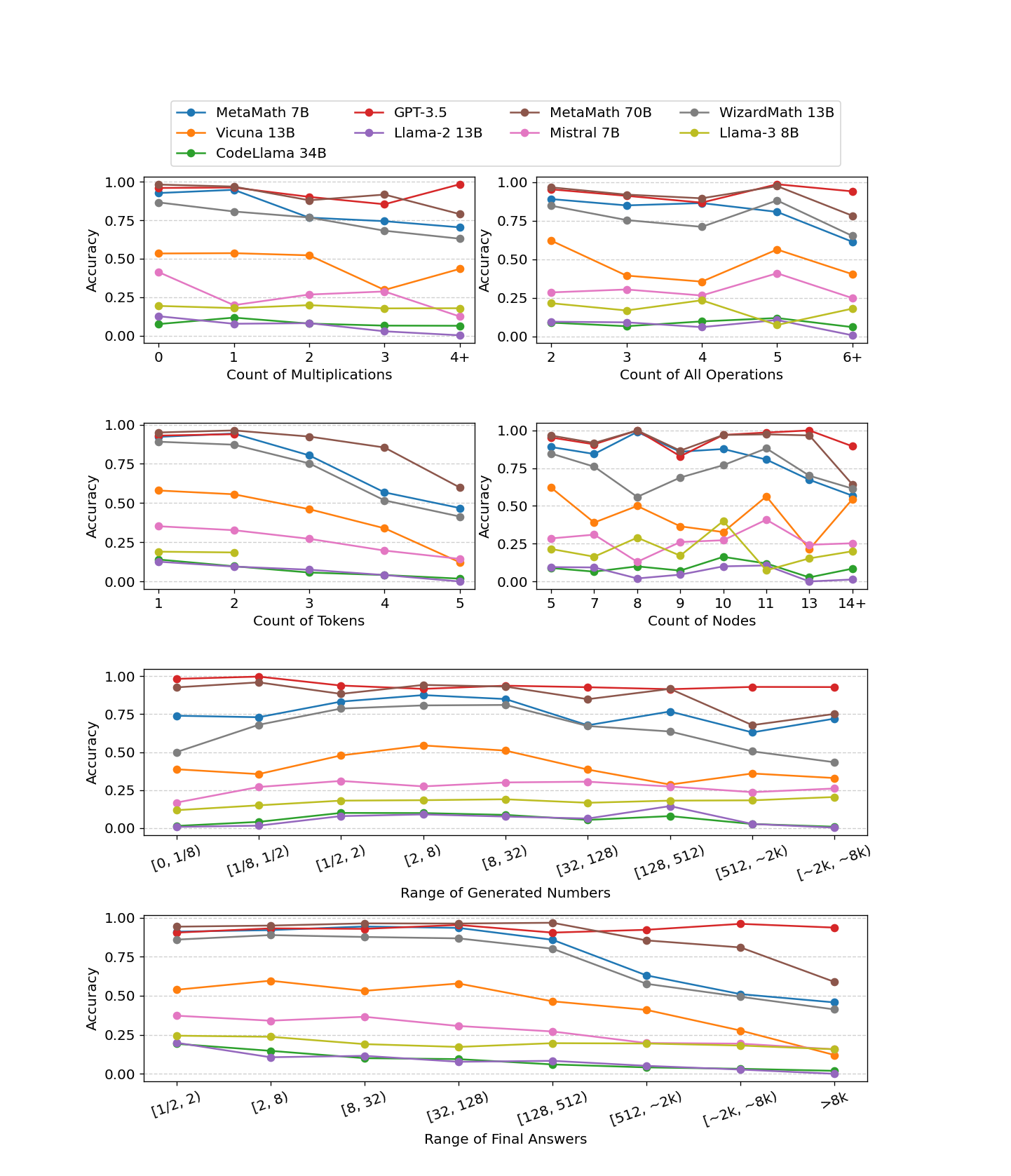}
    \caption{The accuracy of LLMs is plotted against five relevant features when tasked with all questions generated by $M3$. The \textbf{top left} graph shows the accuracy of models against problems with various counts of multiplications, and the \textbf{top right} graph compares problems with various counts of all operations, including both binary and unary operations. The \textbf{middle left} graph shows the accuracy against token counts of the final answer, and the \textbf{middle right} graph shows the accuracy against the count of nodes in the generated AST. The \textbf{bottom two graphs} shows the accuracy of models with different ranges of generated numbers and the range of final answers.}
    \label{fig:accuracy}
    \vspace{-2em}
\end{figure}


\end{document}